\documentclass{article} 
\usepackage{iclr2026_conference,times}

\iclrfinalcopy

\usepackage{amsthm}
\usepackage{amssymb, amsfonts, amsmath}
\usepackage{enumitem, graphicx}

\usepackage{amsmath,amsfonts,bm}

\def\eqref#1{equation~\ref{#1}}

\def\1{\bm{1}}

\DeclareMathAlphabet{\mathsfit}{\encodingdefault}{\sfdefault}{m}{sl}
\SetMathAlphabet{\mathsfit}{bold}{\encodingdefault}{\sfdefault}{bx}{n}

\newcommand{\E}{\mathbb{E}}

\usepackage{subcaption}   
\usepackage{tikz} 
\usetikzlibrary{positioning,calc,arrows,arrows.meta,bending} 

\tikzset{
  obs/.style={circle, draw, thick, minimum size=18pt, inner sep=0pt, font=\small, fill=white},
  noise/.style={rectangle, draw, thick, rounded corners=2pt, minimum width=18pt, minimum height=12pt, inner sep=1pt, font=\scriptsize, fill=white},
  lat/.style={circle, draw, thick, dashed, minimum size=18pt, inner sep=0pt, font=\small, fill=white},
  dir/.style={-{Latex[length=2.4mm]}, line width=0.9pt}, 
  bidi/.style={<->, >=Latex, line width=0.9pt},         
  lbl/.style={font=\scriptsize, fill=white, inner sep=1pt}
}

\usepackage{hyperref}
\usepackage{url}
\usepackage{algorithm}      
\usepackage{algpseudocode}

\renewcommand{\1}{\mathbf{1}}
\newcommand{\0}{\mathbf{0}}
\newcommand{\I}{\mathbf{I}}

\newcommand{\tr}{\mathrm{tr}}
\newcommand{\rank}{\mathrm{rank}}
\newcommand{\diag}{\mathrm{diag}}

\newcommand{\X}{\mathbf{X}}
\newcommand{\x}{\mathbf{x}}

\newcommand{\B}{\mathbf{B}}
\renewcommand{\S}{\mathbf{S}}
\newcommand{\OmegaMat}{\bm{\Omega}}
\newcommand{\SigmaMat}{\bm{\Sigma}}

\newcommand{\T}{\mathbf{T}}        
\newcommand{\Lmat}{\mathbf{L}}     

\theoremstyle{plain}
\newtheorem{theorem}{Theorem}[section]
\newtheorem{lemma}[theorem]{Lemma}

\theoremstyle{definition}

\newtheorem{assumption}[theorem]{Assumption}

\title{DAG DECORation: \\Continuous Optimization for Structure Learning under Hidden Confounding}

\author{
Samhita Pal \\
Department of Biostatistics\\
Vanderbilt University Medical Center\\
Nashville, TN 37232, USA \\
\texttt{samhita.pal@vumc.org} \\
\And
James O'quinn \\
Department of Biostatistics\\
Vanderbilt University\\
Nashville, TN 37235, USA \\
\texttt{james.m.oquinn@vanderbilt.edu} \\
\And
Kaveh Aryan \\
Department of Informatics\\
King's College London\\
London WC2R 2LS, UK \\
\texttt{kaveh.aryan@kcl.ac.uk} \\
\And
Heather Pua \\
Dpt. of Pathology, Microbiology and Immunology\\
Vanderbilt University Medical Center\\
Nashville, TN 37232, USA \\
\texttt{heather.pua@vumc.org} \\
\And
James P. Long \\
Department of Biostatistics\\
MD Anderson Cancer Center\\
Houston, TX 77030, USA \\
\texttt{jplong@mdanderson.org} \\
\And
Amir Asiaee \\
Department of Biostatistics\\
Vanderbilt University Medical Center\\
Nashville, TN 37232, USA \\
\texttt{amir.asiaeetaheri@vumc.org} \\
}

\begin{document}

\maketitle

\begin{abstract}
We study structure learning for linear Gaussian SEMs in the presence of latent confounding. Existing continuous methods excel when errors are independent, while deconfounding-first pipelines rely on pervasive factor structure or nonlinearity. We propose \textsc{DECOR}, a single likelihood-based and fully differentiable estimator that jointly learns a DAG and a correlated noise model. Our theory gives simple sufficient conditions for global parameter identifiability: if the mixed graph is bow free and the noise covariance has a uniform eigenvalue margin, then the map from $(\B,\OmegaMat)$ to the observational covariance is injective, so both the directed structure and the noise are uniquely determined. The estimator alternates a smooth-acyclic graph update with a convex noise update and can include a light bow complementarity penalty or a post hoc reconciliation step. On synthetic benchmarks that vary confounding density, graph density, latent rank, and dimension with $n<p$, \textsc{DECOR} matches or outperforms strong baselines and is especially robust when confounding is non-pervasive, while remaining competitive under pervasiveness. 
\end{abstract}

\section{Introduction}
Directed graphical models, especially directed acyclic graphs (DAGs), provide a powerful formalism for representing causal relationships among variables in domains such as biology, economics, and the social sciences~\citep{pearl2009causality, spirtes2000causation}. However, learning the underlying DAG structure from purely observational data remains a fundamental challenge. Even under the linear Gaussian structural equation model (SEM), the observational distribution is generally consistent with an entire Markov equivalence class of DAGs—distinct graphs that encode the same set of conditional independencies~\citep{chickering2002learning, andersson1997characterization}. Consequently, without further assumptions or interventional data, the true causal structure is unidentifiable~\citep{squires2023causal}.

Although linear Gaussian SEMs are generally identifiable only up to a Markov equivalence class, a notable exception occurs when all error terms share the same variance under causal sufficiency the true DAG is identifiable from purely observational data~\citep{peters2014identifiability}. Outside this equal-variance regime, identifiability typically requires additional asymmetries in the data-generating process, such as non-Gaussian noise (e.g., LiNGAM) or suitable nonlinear additive-noise structure~\citep{shimizu2006linear, hoyer2008nonlinear}. The challenge is further compounded by latent confounders: unobserved variables can induce spurious associations among observed nodes and destroy identifiability for DAGs, shifting the target to partial ancestral or maximal ancestral graphs and PAGs~\citep{richardson2002ancestral, spirtes2000causation, zhang2008causal}. Consequently, learning identifiable causal structure from Gaussian observational data in the presence of latent confounding remains largely open in full generality.

Despite these obstacles, a wave of continuous-optimization methods, initiated by \textsc{NOTEARS}, has advanced DAG discovery from observational data~\citep{zheng2018notears}. These approaches replace the combinatorial acyclicity constraint with a smooth surrogate (e.g., $h(B)=0$ based on a matrix exponential), enabling gradient-based minimization of a likelihood- or score-based objective with sparsity regularization. Follow-ups extend the template to various directions ~\citep{zheng2020learning,yu2019daggnn,lachapelle2019gradient,brouillard2020differentiable,bello2022dagma}. Likelihood-centric variants, such as \textsc{GOLEM}, make the connection explicit by optimizing the Gaussian (equal- or non-equal-variance) log-likelihood under the smooth acyclicity constraint~\citep{ng2020role}. Across this family, a common assumption is causal sufficiency (no unmeasured confounding) with mutually independent noise terms; in practice, violations of this assumption, e.g., latent confounders—can bias edge orientation and degrade recovery~\citep{spirtes2000causation}.

Complementary progress on handling hidden confounding has emerged along two fronts. 
First, methods that exploit distributional asymmetries in non-Gaussian models build on the LiNGAM paradigm~\citep{shimizu2006linear}. A particularly useful structural assumption in this regime is \emph{bow-freeness}, which forbids any unordered pair of observed variables from carrying both a directed edge and a bidirected error link. Bow-free constraints yield identifiability results for mixed graphs in the non-Gaussian setting, and recent work leverages this to orient edges and detect latent siblings without prior knowledge of the number or placement of confounders~\citep{wang2023bang}. Related results establish parameter identifiability—of edge coefficients and noise covariances—under linear Gaussian SEMs on acyclic mixed graphs, including generalized bow-free structures; in particular, \citet{drton2011global}.

Second, deconfounding-first strategies estimate latent influences before DAG discovery proceeds. In this pipeline, one first recovers low-dimensional latent structure from observational data—using factor or spectral methods, principal components, or low-rank plus sparse decompositions—then removes the estimated confounding signal prior to causal graph learning~\citep{frot2019robust,shah2020,agrawal2023decamfounder,squires2022latent,chandrasekaran2010latent}. These approaches typically rely on a \emph{pervasive confounding} assumption, namely that a small number of latent factors load on many observed variables with non-negligible strength, which makes the confounding component identifiable by PCA-type estimators.

Despite this progress, a gap remains. To our knowledge, no continuous-optimization approach both removes latent confounding and learns the DAG in linear Gaussian SEMs when confounding is non-pervasive, that is, when latent factors do not load broadly across many variables. Existing smooth-acyclicity methods typically assume causal sufficiency, and deconfounding-first pipelines rely on pervasive factor structure for identifiability. 


\subsection{Our contribution}
We introduce \textsc{DECOR} (\textbf{DE}confounding via \textbf{CO}rrelation \textbf{R}emoval), a single, differentiable, score-based procedure for learning linear Gaussian SEMs with latent confounding. \textsc{DECOR} departs from two-stage pipelines by modeling correlated noise directly and optimizing a likelihood-aligned score under a smooth acyclicity constraint $h(B)=0$ with sparsity regularization. Our contributions are:

\begin{enumerate}[leftmargin=*, itemsep=2pt, topsep=2pt]
\item \textbf{Identifiability under bow-free structure.} We establish sufficient conditions for global parameter identifiability in linear Gaussian SEMs with correlated errors. If the directed mixed graph is bow-free and the error covariance has a uniform eigenvalue margin, then the model satisfies generalized bow-freeness~\citep{drton2011global}. Consequently, the parametrization that maps a weighted adjacency matrix and an error covariance to the observational covariance is injective, so with sufficient samples to estimate the data covariance accurately, both the causal structure and the noise covariance are uniquely recoverable. This covers both pervasive and non-pervasive confounding and yields uniqueness of structure and noise parameters from observational data.

\item \textbf{A continuous and differentiable DAG estimator in the presence of latent confounding.} We develop a likelihood-based continuous optimization framework that jointly estimates the DAG and a structured error covariance without requiring pervasiveness. The procedure alternates between two steps: updating the graph under a smooth acyclicity constraint with sparsity regularization, and updating the noise covariance within a stable parametrization that maintains a fixed eigenvalue margin. This blockwise design is modular, so one can pair any gradient-based optimizer for the graph step with any compatible covariance estimator for the noise step while respecting the required constraints.

\item \textbf{Integrated deconfounding and discovery.} \textsc{DECOR} replaces the usual deconfounding-then-DAG pipeline with a single estimator that removes latent correlations while orienting edges. Under our identifiability conditions, this yields consistent structure recovery and improves robustness when confounding is sparse or localized rather than pervasive.

\item \textbf{Empirical validation.} Across synthetic and real benchmarks, \textsc{DECOR} matches or outperforms strong baselines from smooth-acyclicity methods, classic constraint and score-based approaches designed to handle latent variables, and deconfounding-first pipelines, over a range of confounding regimes.
\end{enumerate}


As for the structure of the paper, we first formalize the problem and review background on linear Gaussian SEMs with latent confounding. Next, we introduce the \textsc{DECOR} framework, state our identifiability results, and describe the optimization procedure. We then present empirical evaluations on synthetic benchmarks and compare against strong baselines, highlighting where \textsc{DECOR} offers practical advantages.

\subsection{Problem formulation}

We consider $p$ observed variables indexed by $V=\{1,\dots,p\}$ generated by a linear Gaussian SEM with possibly correlated errors:
\begin{equation}
\label{eq:sem}
\x \;=\; \B^{\top}\x \;+\; \mathbf{e}, 
\qquad \mathbf{e}\sim\mathcal{N}(\0,\OmegaMat), 
\qquad \text{acyclicity: } h(\B)=0,
\end{equation}
where $\B\in\mathbb{R}^{p\times p}$ is the weighted adjacency matrix of a DAG, $\OmegaMat\succ0$ is the noise covariance, and $h(\cdot)$ is a smooth surrogate that enforces acyclicity. The implied covariance and precision are
\begin{equation}
\label{eq:sigma-theta}
\SigmaMat \;=\; (\I-\B)^{-1}\,\OmegaMat\,(\I-\B)^{-\top}, 
\qquad 
\SigmaMat^{-1} \;=\; (\I-\B)\,\OmegaMat^{-1}\,(\I-\B)^{\top}.
\end{equation}

Given $n$ i.i.d. samples $\x_{1},\ldots,\x_{n}\sim\mathcal{N}(\0,\SigmaMat)$ arranged as rows of $\X\in\mathbb{R}^{n\times p}$, a negative log-likelihood for $(\B,\OmegaMat)$, up to additive constants, is
\begin{equation}
\label{eq:nll}
\mathcal L_n(\B,\OmegaMat) 
\;=\; \frac{1}{n}\,\bigl\lVert \OmegaMat^{-1/2}\bigl(\X-\X\B\bigr)\bigr\rVert_{F}^{2} 
\;+\; \log\det\OmegaMat 
\;-\; 2 \log\det(\I-\B).
\end{equation}

\paragraph{Connections to existing objectives.}
(i) If $\OmegaMat=\I$ and $h(\B)=0$ enforces a DAG, then after a topological ordering $\det(\I-\B)=1$, so $\mathcal L_n$ reduces to least squares on the residuals plus a constant. With sparsity regularization on $\B$, this recovers the \textsc{NOTEARS} objective \citep{zheng2018notears}. 
(ii) The \textsc{GOLEM} family optimizes the Gaussian likelihood \emph{without} a separate smooth acyclicity penalty. Then convert the last term to a regularizer as $-2\log\bigl|\det(\I-\B)\bigr|$, which equals zero for DAGs\citep{ng2020role}.

\paragraph{Mixed-graph notation.}
For node $i$, let $P(i)$ be the set of directed parents of $i$, and let $S(i)$ be the set of nodes that share a bidirected edge with $i$. For any matrix $M$, $M_{R,C}$ denotes the submatrix with rows $R$ and columns $C$. We write $[i] = \{1,\ldots,i\}$.

\section{Related Work}

\subsection{DAG Discovery via Continuous Optimization}

Classical approaches to causal discovery include constraint-based methods such as 
PC and FCI \citep{spirtes2000} and score-based procedures like GES \citep{chickering2002ges}. Current researches have also proposed computationally faster constraint-based causal discovery methods \citep{colombo2012, bernstein2020ordering, shiragur2024causal, pal2025penalized}.
More recently, continuous optimization has emerged as a powerful alternative. 
NOTEARS \citep{zheng2018notears} introduced a differentiable acyclicity constraint, 
allowing gradient-based optimization to recover sparse DAGs. 
Follow-up work refined this paradigm through alternative characterizations of acyclicity 
\citep{bello2022dagma}, nonlinear extensions \citep{yu2019daggnn}, and sparsity-regularized 
likelihoods such as GOLEM \citep{ng2020role}. 

Despite their success, these methods generally recover graphs only up to Markov equivalence 
and assume causal sufficiency. Concerns have also been raised about spurious optima 
and reliance on data-specific artifacts \citep{reisach2021beware,seng2023harder}. 
Recent results address these issues by showing that carefully regularized scores can recover 
the sparsest representative of the equivalence class under mild conditions \citep{deng2024markov}, 
but most approaches remain limited to the confounder-free case.

\subsection{Deconfounding in Causal Discovery}
The second line adopts a \emph{deconfounding-first} strategy: estimate latent structure, remove its effect, then learn a DAG on residuals. Concretely, one may recover a low-rank confounding component alongside a sparse conditional graph~\citep{frot2019robust,shah2020}, fit approximate factor models under pervasiveness to extract latent scores~\citep{wang2019blessings,squires2022latent}, or use spectral summaries to enable downstream edge orientation, as in DeCAMFounder~\citep{agrawal2023decamfounder}. These pipelines are computationally attractive and work well when a few latent factors influence many observables, yet their guarantees typically hinge on pervasiveness or nonlinearity and thus do not yield global identifiability for linear Gaussian SEMs with possibly non-pervasive confounding. Moreover, they usually target only the causal graph, treating the noise covariance as a nuisance; the confounding component is estimated and subtracted rather than modeled and identified.

To distinguish our contribution, we briefly detail two recent deconfounder methods.

\paragraph{Low-rank plus sparse precision decomposition.}
\citet{frot2019robust} assume that the observed precision matrix decomposes into a sparse component that encodes conditional relations among observables and a low-rank component induced by a small number of latent factors with pervasive loadings. Under compatibility or incoherence conditions that prevent the low-rank part from mimicking sparsity \citep{chandrasekaran2010latent}, together with appropriate sample-size and tuning regimes, this split is identifiable. Intuitively, few hidden variables must influence many measured variables, while the conditional graph among observables remains genuinely sparse. 

\paragraph{DeCAMfounder: deconfounding via additive-noise identifiability.}
\citet{agrawal2023decamfounder} target identifiability by first summarizing pervasive confounding through estimated sufficient statistics of a latent factor, then orienting edges among observables using additive-noise identifiability. Concretely, the method fits nonlinear parental mechanisms with smoothness assumptions and Gaussian disturbances conditional on the confounder summary; under these functional and distributional restrictions, the causal ordering among observed variables is identifiable from the conditional law. In purely linear-Gaussian regimes, by contrast, one typically recovers only a Markov equivalence class, so nonlinearity is essential for identification in this approach. 

\noindent Our work differs in both scope and assumptions: we remain in the linear Gaussian setting, allow correlated errors induced by possibly non-pervasive confounding, and obtain global parameter identifiability under bow-free structure with a uniform eigenvalue margin on the noise covariance, leading to a single continuous optimization procedure that jointly estimates the directed structure and correlated noise.

\section{Global identifiability related to continuous optimization}
We summarize the graphical and algebraic identifiability results of \citet{drton2011global} in the linear Gaussian SEM on an acyclic mixed graph, which forms the basis of our contribution.

\paragraph{Graphical intuition.}
In a mixed graph with directed edges encoded by $\B$ and bidirected edges by $\OmegaMat$, consider any induced subset of nodes and examine its two layers of edges: the directed part and the bidirected part (capturing error correlations from latent confounding). A fundamental obstruction to identifiability arises precisely when the directed edges on that subset form a \emph{converging arborescence}, a directed tree in which every node has a unique directed path into a single \emph{sink}, and, simultaneously, the bidirected edges on the same nodes form a connected graph. Intuitively, the sink aggregates all upstream total effects, and if the same nodes are fully tied together by confounding, then directed influence and correlated noise become inseparable at the level of second moments. This pattern strictly generalizes \emph{bow-freeness}: for two nodes, a directed edge together with a bidirected edge is exactly a bow; for larger subsets, the ``generalized bow'' is an in-arborescence to one sink overlaid with a connected bidirected component. Consequently, global identifiability is equivalent to the \emph{absence} of any induced subset exhibiting this obstruction.

\paragraph{Algebraic rank form.}
The same condition can be expressed as a clean nodewise rank requirement that couples a block of the noise matrix with the upstream effect map. Let $\T=(\I-\B)^{-1}$ be the total effect matrix. For node $i$, let $P(i)$ be its set of directed parents and $S(i)$ the set of bidirected neighbors. Then global identifiability is equivalent to the following full column rank condition at every non-sink node:
\begin{equation}
\label{eq:rank-cond}
\rank\!\bigl(\,\underbrace{\OmegaMat_{[i]\setminus S(i),\,[i]}}_{\text{noise block}} \;\;\underbrace{\T_{[i],\,P(i)}}_{\text{effect block}}\,\bigr) \;=\; |P(i)| \quad \text{for all } i\in\{1,\ldots,p-1\}.
\end{equation}
This algebraic test rules out precisely the edge patterns that confound directed influence with correlated errors, and it does so without invoking low-rank plus sparse decompositions, incoherence bounds, or nonlinear additive-noise assumptions.

\paragraph{Injectivity.}
The covariance parametrization maps parameters to the observational covariance via
\[
(\B,\OmegaMat)\;\longmapsto\;\SigmaMat \;=\; (\I-\B)^{-1}\,\OmegaMat\,(\I-\B)^{-\top}.
\]
Global identifiability means this map is \emph{injective}: if two parameter pairs $(\B,\OmegaMat)$ and $(\B',\OmegaMat')$ produce the same $\SigmaMat$, then $(\B,\OmegaMat)=(\B',\OmegaMat')$. Under the graphical or rank conditions above, injectivity holds, so both the edge weights and the noise covariance are uniquely determined by the observational covariance  \citep{drton2011global}.

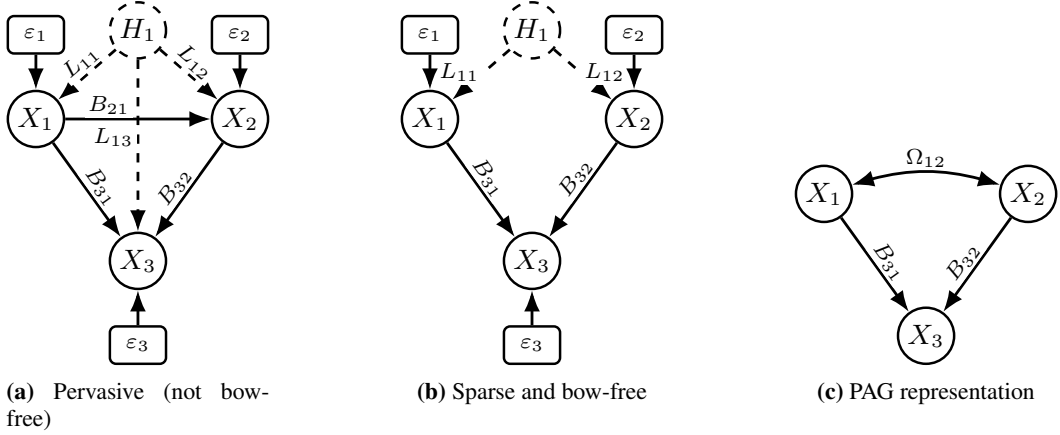
\begin{figure}[t]
\centering

\begin{subfigure}[t]{0.25\textwidth}
\centering
\resizebox{\linewidth}{!}{%
\begin{tikzpicture}[scale=1]
  \node[obs] (A1) at (0,0) {$X_1$};
  \node[obs] (A2) at (2.3,0) {$X_2$};
  \node[obs] (A3) at (1.15,-1.6) {$X_3$};

  \node[noise] (Ae1) at (0,0.95) {$\varepsilon_1$};
  \node[noise] (Ae2) at (2.3,0.95) {$\varepsilon_2$};
  \node[noise] (Ae3) at (1.15,-2.55) {$\varepsilon_3$};

  \node[lat] (AL) at (1.15,1.0) {$H_{1}$};
  \draw[dir, dashed] (AL) -- node[lbl, sloped,above, pos=.45] {$L_{11}$}(A1);
  \draw[dir, dashed] (AL) -- node[lbl, above, sloped, pos=.45] {$L_{12}$}(A2);
  \draw[dir, dashed] (AL) -- node[lbl, left, pos=.45] {$L_{13}$}(A3);

  \draw[dir] (A1) -- node[lbl, above, pos=.30] {$B_{21}$} (A2);
  \draw[dir] (A1) -- node[lbl, sloped, above, pos=.55] {$B_{31}$} (A3);
  \draw[dir] (A2) -- node[lbl, sloped, above, pos=.55] {$B_{32}$} (A3);

  \draw[dir] (Ae1) -- (A1);
  \draw[dir] (Ae2) -- (A2);
  \draw[dir] (Ae3) -- (A3);
\end{tikzpicture}%
}
\caption{Pervasive (not bow-free)}
\end{subfigure}
\hfill
\begin{subfigure}[t]{0.25\textwidth}
\centering
\resizebox{\linewidth}{!}{%
\begin{tikzpicture}[scale=1]
  \node[obs] (B1) at (0,0) {$X_1$};
  \node[obs] (B2) at (2.3,0) {$X_2$};
  \node[obs] (B3) at (1.15,-1.6) {$X_3$};

  \node[noise] (Be1) at (0,0.95) {$\varepsilon_1$};
  \node[noise] (Be2) at (2.3,0.95) {$\varepsilon_2$};
  \node[noise] (Be3) at (1.15,-2.55) {$\varepsilon_3$};

  \node[lat] (BL) at (1.15,1.0) {$H_{1}$};
  \draw[dir, dashed] (BL) -- node[lbl, left, pos=.45] {$L_{11}$}(B1);
  \draw[dir, dashed] (BL) -- node[lbl, right, pos=.45] {$L_{12}$}(B2);

  \draw[dir] (B1) -- node[lbl, above, sloped,pos=.45] {$B_{31}$} (B3);
  \draw[dir] (B2) -- node[lbl, sloped, above, pos=.45] {$B_{32}$} (B3);

  \draw[dir] (Be1) -- (B1);
  \draw[dir] (Be2) -- (B2);
  \draw[dir] (Be3) -- (B3);
\end{tikzpicture}%
}
\caption{Sparse and bow-free}
\end{subfigure}
\hfill
\begin{subfigure}[t]{0.25\textwidth}
\centering
\resizebox{\linewidth}{!}{%
\begin{tikzpicture}[scale=1]
  \node[obs] (P1) at (0,0) {$X_1$};
  \node[obs] (P2) at (2.3,0) {$X_2$};
  \node[obs] (P3) at (1.15,-1.6) {$X_3$};

  \draw[bidi, bend left=18] (P1) to node[lbl, above] {$\Omega_{12}$} (P2);
  \draw[dir] (P1) -- node[lbl, sloped, above, pos=.55] {$B_{31}$} (P3);
  \draw[dir] (P2) -- node[lbl, sloped, above, pos=.55] {$B_{32}$} (P3);
\end{tikzpicture}%
}
\caption{PAG representation}
\end{subfigure}

\caption{Pervasive versus bow-free structures and their PAG summary. Latent-to-\(X_i\) edges are labeled with \(L_{1i}\).}
\end{figure}

\subsection{A simple sufficient characterization for global identifiability}

We give a new, checkable route to the nodewise rank condition that underpins global identifiability of linear Gaussian SEMs on acyclic mixed graphs. The idea is purely structural and numerical: rule out local bows in the graph, and keep a uniform positive margin away from singularity in the noise. Together these two ingredients force the rank tests to pass at every node, which in turn yields injectivity of the covariance map \((\B,\OmegaMat)\mapsto\SigmaMat\).

We now give simple sufficient assumptions that guarantee the rank conditions defined in Equation \eqref{eq:rank-cond} hold without having to inspect all induced subgraphs:

\begin{assumption}[Bow-free]\label{asmp:bow}
For every node \(i\), the parent set and the sibling set are disjoint: \(P(i)\cap S(i)=\varnothing\), i.e., in the linear Gaussian SEM $\forall i, j: B_{ij} \Omega_{ij} = 0$. 
\end{assumption}

\begin{assumption}[Eigenvalue margin]\label{asmp:eig}
The noise covariance is uniformly well conditioned: \(\OmegaMat\succ0\) and \(\lambda_{\min}(\OmegaMat)\ge \varepsilon>0\).
\end{assumption}
\paragraph{Intuition.}
The nodewise rank test combines two ingredients: an \emph{effect block}, which carries the directed influence of a node’s parents into the node, and a \emph{noise block}, which carries correlations induced by latent confounders. For a given node \(i\), let \(P(i)\) be its directed parents and \(S(i)\) its bidirected neighbors (siblings). In the rank test we keep rows that are informative about \(i\)’s directed inputs, and we drop rows indexed by \(S(i)\) because those rows are contaminated by the same confounding that also touches \(i\). Assumption~\ref{asmp:bow} (bow-free) guarantees that none of the dropped rows belongs to a parent, so we do not accidentally remove parent information. Assumption~\ref{asmp:eig} (eigenvalue margin) ensures that the remaining rows of the noise block are well conditioned, so they cannot numerically cancel the clear ``parent signatures'' present in the effect block. 

To make this precise, recall that \(\T=(\I-\B)^{-1}\) is the total effect map. In a topological order of the DAG, \(\T\) is unit lower triangular. Hence the submatrix of \(\T\) that collects columns for \(P(i)\) and rows up to \(i\) embeds an identity on the parent rows. These identity columns are the parent signatures. The noise block multiplies these signatures. If the noise block is full row rank with a positive singular value margin, it cannot eliminate those signatures. The product therefore has as many independent columns as there are parents, which is exactly the nodewise rank condition.

\begin{lemma}[Effect block is full column rank]\label{lem:effect}
Under acyclicity, reorder variables in a topological order so that \(\T\) is unit lower triangular. Then, for any node \(i\), the submatrix \(\T_{[i],\,P(i)}\) has full column rank, and the rows indexed by \(P(i)\) contain an identity on the parent columns.
\end{lemma}

\begin{lemma}[Noise block retains a margin]\label{lem:noise}
Under Assumption~\ref{asmp:eig}, for every node \(i\) the rectangular block \(\OmegaMat_{[i]\setminus S(i),\,[i]}\) has full row rank. In particular, its smallest nonzero singular value is bounded below by a positive constant that depends only on the eigenvalue margin.
\end{lemma}

\begin{theorem}[Deterministic identifiability under a bow and a margin]\label{thm:ident}
Assume acyclicity, Assumption~\ref{asmp:bow}, and Assumption~\ref{asmp:eig}. Then, for every node \(i\),
\[
\rank\!\Bigl(\OmegaMat_{[i]\setminus S(i),\,[i]}\;\T_{[i],\,P(i)}\Bigr)=|P(i)|.
\]
Consequently, the covariance parametrization is injective:
\[
(\B,\OmegaMat)\;\mapsto\;\SigmaMat=(\I-\B)^{-1}\,\OmegaMat\,(\I-\B)^{-\top}\quad\text{is one to one.}
\]
Hence both the edge weights and the noise covariance are uniquely determined by the observational covariance.
\end{theorem}

\paragraph{Remarks.}
\begin{itemize}[leftmargin=*, itemsep=2pt, topsep=2pt]
\item Bow-freeness alone prevents the simplest graphical obstruction but does not guarantee identifiability. The eigenvalue margin supplies a quantitative separation so that parent signatures in the effect block cannot be washed out by the noise block.
\item The conditions do not assume pervasiveness. They allow non-pervasive confounding patterns, since no factor model or low-rank recovery is required.
\item Graphically, bow-freeness is the two-node special case of the more general obstruction where a directed in-arborescence into a single sink is tied together by bidirected edges. Our conditions avoid this obstruction without scanning all induced subsets.
\item The result is deterministic. Statistical consistency follows once the sample covariance concentrates near \(\SigmaMat\) and the estimator targets the likelihood under these constraints.
\end{itemize}


While bow-freeness is not, by itself, an identifiability guarantee, encouraging it during estimation improves well-posedness and interpretability. We therefore propose to add a soft \emph{complementarity} regularizer that discourages a directed edge and residual correlation on the same unordered pair:
\begin{equation}
\label{eq:bow-pen}
\Phi_{\text{bow}}(\B,\OmegaMat)
\;=\;
\sum_{i<j} \omega_{ij}\,\bigl(|B_{ij}|+|B_{ji}|\bigr)\,|\Omega_{ij}|,
\qquad \omega_{ij}\ge 0,
\end{equation}
with weights \(\omega_{ij}\) chosen as constants or data-adaptive scores. This symmetrized product penalizes any causal channel between \(i\) and \(j\) that coexists with residual correlation in \(\OmegaMat_{ij}\), which nudges the estimator toward bow-free patterns that are friendlier to the rank test.

\section{Proposed DECOR Method}
Let \(\X\in\mathbb{R}^{n\times p}\) be the data matrix whose rows are i.i.d.\ samples from the linear structural equation model (SEM) in \eqref{eq:sem}, where \(\B\in\mathbb{R}^{p\times p}\) encodes directed effects (edge \(j\!\to\! i\) iff \(\B_{ij}\neq 0\)), \(\bm\Omega\succ 0\) is the error covariance, and \(\bm\Theta:=\bm\Omega^{-1}\) is the precision. Define the residual matrix \(\mathbf{E}(\B):=\X-\X\B\) and the residual covariance \(\widehat S_\mathbf{E}(\B):=n^{-1}\mathbf{E}(\B)^{\top}\mathbf{E}(\B)\). Acyclicity is enforced by the differentiable NOTEARS surrogate \(h(\B)=\mathrm{tr}(\exp(\B\circ \B))-p\), where \(\circ\) denotes the Hadamard product.

\paragraph{Objective and constraints.}
We estimate the directed matrix \(B\in\mathbb{R}^{p\times p}\) and the noise covariance \(\Omega\succ0\) by minimizing a residual-likelihood with sparsity, acyclicity, and bow-freeness control:
\begin{align*}
    &\min_{\B,\ \bm\Omega\succ0}\;\Big\{
\underbrace{\tfrac{1}{n}\,\tr\!\big((\X-\X\B)^\top\bm\Omega^{-1}(\X-\X\B)\big)}_{\text{residual likelihood}}
\;+\;\underbrace{\log\det\bm\Omega}_{\text{normalizer}}
\;+\;\lambda_\B\|\B\|_1
\;+\;\lambda_{\bm\Omega}\|\bm\Omega_{\mathrm{off}}\|_1\\
& \qquad \qquad \;+\;\lambda_{\mathrm{bow}}\!\sum_{i<j}\omega_{ij}\,|B_{ij}|\,|\Omega_{ij}| \Big\}
\quad\text{s.t. } h(\B)=0.
\end{align*}
Here \(\X\in\mathbb{R}^{n\times p}\) is the data, \(\|\cdot\|_1\) is the entrywise \(\ell_1\) norm that promotes sparsity, \(\|\bm\Omega_{\mathrm{off}}\|_1\) penalizes off-diagonal entries of \(\Omega\), \(h(\B)=\tr(\exp(\B\circ \B))-p\) is the NOTEARS acyclicity surrogate, and \(\sum_{i<j}\omega_{ij}|B_{ij}||\Omega_{ij}|\) is the soft bow-freeness penalty that discourages simultaneous directed and bidirected links on the same pair. Writing \(\bm\Theta=\bm\Omega^{-1}\) and \(\mathbf{E}(\B)=\X-\X\B\), the loss term equals \(\tfrac{1}{n}\|\,\bm\Theta^{1/2}\mathbf{E}(\B)\,\|_F^2\).

\paragraph{Biconvex core without acyclicity and bow.}
Without \(h(\B)=0\) and the bow term, the problem
\[
\min_{\B,\ \bm\Omega\succ0}\;
\frac{1}{n}\,\tr\!\big(\mathbf{E}(\B)^\top\bm\Omega^{-1}\mathbf{E}(\B)\big)+\log\det\bm\Omega
+\lambda_\B\|\B\|_1+\lambda_{\bm\Omega}\|\bm\Omega_{\mathrm{off}}\|_1
\]
is \emph{biconvex}: for fixed \(\bm\Omega\) (equivalently fixed \(\bm\Theta\)), the objective in \(\B\) is a convex quadratic plus \(\ell_1\) term since \(n^{-1}\|\,\bm\Theta^{1/2}(\X-\X\B)\,\|_F^2\) is convex in \(\B\); for fixed \(\B\), the optimization over \(\bm\Omega\) (covariance route) or over \(\bm\Theta\) (precision route) is convex (graphical lasso–type). Alternating minimization is therefore natural and leads to a stationary point of this biconvex core.

\paragraph{Effect of acyclicity and bow penalties.}
Reintroducing the acyclicity constraint \(h(\B)=0\) renders the \(\B\)-subproblem \emph{nonconvex}; NOTEARS handles this with a smooth equality surrogate inside an augmented-Lagrangian proximal-gradient scheme. Adding the bow penalty further couples the blocks nonlinearly and nonsmoothly through \(|\B_{ij}||\bm\Omega_{ij}|\), making each subproblem harder and the overall landscape more intricate.

\paragraph{Practical enforcement of bow-freeness.}
Instead of optimizing with the explicit bow penalty (which slows and complicates Stage~1), we adopt a post-hoc, computationally light enforcement that preserves the favorable biconvex structure during optimization. After alternating between the \(\B\)-update and the noise update to convergence, we apply hard thresholding and a one-per-pair reconciliation:
\[
\widehat B^{\,\mathrm{thr}}_{ij}=\widehat B_{ij}\,\mathbf{1}\{|\widehat B_{ij}|\ge\tau_\B\},\qquad
\widehat{\Omega}^{\,\mathrm{thr}}_{ij}=\widehat{\Omega}_{ij}\,\mathbf{1}\{i\neq j,\ |\widehat{\Omega}_{ij}|\ge\tau_{\bm\Omega}\},
\]
followed by, for each unordered pair \(\{i,j\}\) with both a directed edge (either \(\widehat B^{\,\mathrm{thr}}_{ij}\) or \(\widehat B^{\,\mathrm{thr}}_{ji}\)) and a bidirected edge (\(\widehat{\Omega}^{\,\mathrm{thr}}_{ij}\)) remaining, keeping the stronger channel and zeroing the other, e.g.
\(
\text{if }\ \max\{|\widehat B^{\,\mathrm{thr}}_{ij}|,|\widehat B^{\,\mathrm{thr}}_{ji}|\}
\;\ge\; c\cdot {|\widehat{\Omega}^{\,\mathrm{thr}}_{ij}|}/{\sqrt{\widehat{\Omega}^{\,\mathrm{thr}}_{ii}\,\widehat{\Omega}^{\,\mathrm{thr}}_{jj}}}\
\) keep the directed edge, else keep the bidirected edge.
This strategy retains convex subproblems during the alternation (fast and stable), avoids coupling biases in the \(\B\)-step, and empirically improves precision with minimal recall loss while exactly enforcing bow-freeness in the final graph. In short, alternating on the biconvex core and enforcing bow-freeness by post-hoc hard thresholding is a pragmatic and effective solution to an otherwise nonconvex, tightly coupled optimization. 

\subsection{Stage~1: directed part (NOTEARS-style)}
Given a current weight \(\bm\Theta\succ 0\), estimate \(\B\) by minimizing the residual-weighted objective under acyclicity,
\begin{equation}
\label{eq:s1}
\min_{B}\;\; \frac{1}{n}\,\mathrm{tr}\!\big(\mathbf{E}(\B)^{\top}\bm\Theta\,\mathbf{E}(\B)\big)\;+\;\lambda_\B\|\B\|_{1}
\quad \text{s.t. } h(\B)=0.
\end{equation}
The gradient of the smooth part is
\(
\nabla_\B\!\left[\tfrac{1}{n}\mathrm{tr}\!\big(\mathbf{E}^{\top}\bm\Theta \mathbf{E}\big)\right]
= -\tfrac{2}{n}\,\X^{\top}\bm\Theta\,(\X-\X\B).
\)
Following \cite{zheng2018notears}, we solve Stage~1 by a proximal gradient step on the augmented Lagrangian
$\mathcal L_\rho(\B,\alpha)=n^{-1}\tr\!\big((\X-\X\B)^\top\bm\Theta(\X-\X\B)\big)+\lambda_\B\|\B\|_1+\alpha\,h(\B)+\tfrac{\rho}{2}h(\B)^2$,
where $h(B)=\tr(\exp(B\circ B))-p$ is the differentiable NOTEARS acyclicity surrogate. At iterate $\B$, we take a gradient step on the smooth part and then apply the proximal map of the $\ell_1$ penalty (soft-thresholding), yielding
\[
\B^{+}\leftarrow \operatorname{Soft}_{\eta\lambda_\B}\!\Big(\B-\eta\big[\nabla f(\B;\bm\Theta)+(\alpha+\rho\,h(\B))\,\nabla h(\B)\big]\Big),\qquad \diag(\B^{+})=0,
\]
where $\textstyle \operatorname{Soft}_{\eta\lambda}(Z)=\mathrm{sign}(Z)\cdot\max\!\bigl(|Z|-\eta\lambda,\,0\bigr),
$ with $f(\B;\bm\Theta)=n^{-1}\tr(\E^\top \bm\Theta \E)$ and $\E=\X-\X\B$. The stepsize $\eta$ is chosen by Armijo backtracking to ensure sufficient decrease of $\mathcal L_\rho$, while the augmented-Lagrangian multipliers are updated as
$\alpha\leftarrow \alpha+\rho\,h(\B^{+})$ and $\rho$ is increased when $|h(\B^{+})|$ stalls. Intuitively, the gradient term drives data fit under the current residual weighting $\bm\Theta$, the soft-thresholding induces sparsity in $\B$, and the augmented Lagrangian terms steer the iterate toward acyclicity without hard projection. This is the same mechanism used in NOTEARS (proximal/gradient steps on a smooth objective plus an augmented Lagrangian penalty on $h(\B)$), here with $\bm\Theta$ weighting the residuals. The output of Stage~1 is \(\widehat \B\). It should be noted that any other DAG-learning algorithm could have been used here, instead of NOTEARS.

\subsection{Stage~2: noise part (two interchangeable routes)}
Given \(\widehat \B\), form \(\mathbf{E}=\X-\X\widehat \B\) and \(\widehat S_\mathbf{E}=\tfrac{1}{n}\mathbf{E}^{\top}\mathbf{E}\). Two convex alternatives are used to estimate $\OmegaMat$.

\paragraph{Path~1 (Covariance-route).}
Following \citep{bien2011sparse}, we can optimize \(\bm\Omega\) directly by
\begin{equation}
\label{eq:s2cov}
\min_{\bm\Omega\succ 0}\;\; f_{\mathrm{cov}}(\Omega;\widehat \B)
:= \mathrm{tr}\!\big(\widehat S_\mathbf{E}\,\bm\Omega^{-1}\big) + \log\det\bm\Omega
+ \lambda_{\bm\Omega}\,\|\bm\Omega_{\mathrm{off}}\|_{1}.
\end{equation}
The gradient of the smooth part is \(-\,\bm\Omega^{-1}\widehat S_\mathbf{E}\,\bm\Omega^{-1}+\bm\Omega^{-1}\). A proximal-gradient or proximal-Newton method with soft-thresholding on the off-diagonal entries and an symmetric positive definite (SPD) projection step (eigenvalue flooring or line-search) yields \(\widehat{\bm\Omega}\). In Stage~1, \(\bm\Theta=\bm\Omega^{-1}\) is applied via linear solves (sparse Cholesky or preconditioned conjugate gradient), avoiding explicit inversion.

\paragraph{Path~2 (Precision-route).}
Following \cite{friedman2008sparse}, we can optimize \(\bm\Theta\) by graphical lasso
\begin{equation}
\label{eq:s2prec}
\min_{\bm\Theta\succ 0}\;\; f_{\mathrm{prec}}(\bm\Theta;\widehat \B)
:= \mathrm{tr}\!\big(\widehat S_\mathbf{E}\,\bm\Theta\big) - \log\det \bm\Theta
+ \lambda_{\bm\Theta}\,\|\bm\Theta_{\mathrm{off}}\|_{1}.
\end{equation}
 Coordinate-descent or ADMM solvers produce a sparse \(\widehat{\bm\Theta}\) that can be used directly in \eqref{eq:s1}. If specific \(\bm\Omega_{ij}\) are needed for diagnostics or bow reconciliation, selected entries of \(\bm\Omega=\bm\Theta^{-1}\) can be computed without forming the full inverse by solving \(\bm\Theta v^{(j)}=e_j\) and reading \(\bm\Omega_{ij}=v^{(j)}_i\).

\subsection{Post-hoc bow reconciliation}
After Stage 1 and Stage 2, apply hard thresholding and enforce at most one channel per unordered pair. Concretely, prune small entries in $\widehat{\B}$ and off-diagonals of $\widehat{\OmegaMat}$, then for any pair with both a directed and a bidirected edge, keep the stronger signal and zero the other. Details, including SPD projection for $\widehat{\OmegaMat}$ and acyclicity enforcement for $\widehat{\B}$, appear inside Algorithm~\ref{alg:decor-2s}.


\begin{algorithm}[H]
\small 
\caption{\textsc{DECOR-2S}: unified two-stage estimator with switchable Stage 2}
\label{alg:decor-2s}
\begin{algorithmic}[1]
\State \textbf{Inputs:} data $\X\in\mathbb{R}^{n\times p}$; penalties $(\lambda_{\B},\lambda_{\Omega},\lambda_{\Theta})$; thresholds $(\tau_{\B},\tau_{\Omega})$; route $\in\{\textsc{cov},\textsc{prec}\}$.
\State Compute $\widehat{\S} \gets \tfrac{1}{n}\X^\top\X$ and initialize $\bm{\Theta}^{(0)} \gets \mathrm{diag}(\mathrm{diag}(\widehat{\S}))^{-1}$.
\Statex

\State \textbf{Stage 1: graph update (NOTEARS-style).} \label{line:stage1}
\State Solve \eqref{eq:s1} with current precision $\bm{\Theta}^{(0)}$ (proximal augmented Lagrangian) to obtain $\widehat{\B}$. 
\State Form residuals $\bm E \gets \X - \X\widehat{\B}$ and their covariance $\widehat{\S}_{\bm E} \gets \tfrac{1}{n}\bm E^\top\bm E$.
\Statex

\State \textbf{Stage 2: noise update (switchable).}
\If{$\text{route}=\textsc{cov}$} \label{line:cov-route}
  \State \textbf{Covariance-route:} solve \eqref{eq:s2cov} on $\OmegaMat$ using a proximal SPD solver to get $\widehat{\OmegaMat}$.
  \State Set $\widehat{\bm{\Theta}} \gets \widehat{\OmegaMat}^{-1}$.
\ElsIf{$\text{route}=\textsc{prec}$} \label{line:prec-route}
  \State \textbf{Precision-route:} solve \eqref{eq:s2prec} (graphical lasso on $\widehat{\S}_{\E}$) to get sparse $\widehat{\bm{\Theta}}$.
  \State Set $\widehat{\OmegaMat} \gets \widehat{\bm{\Theta}}^{-1}$.
\EndIf
\Statex

\State \textbf{Post-processing: bow complementarity and thresholds.}
\State Apply complementarity penalty \eqref{eq:bow-pen} or post-hoc reconciliation: for each unordered pair $\{i,j\}$, if $(|\widehat{\B}_{ij}|+|\widehat{\B}_{ji}|)\,|\widehat{\OmegaMat}_{ij}|>\text{tol}$, zero the weaker channel by normalized comparison; enforce SPD on $\widehat{\OmegaMat}$ and acyclicity on $\widehat{\B}$ if needed.
\State Hard-threshold: $\widehat{\B}\leftarrow \mathsf{HT}(\widehat{\B};\tau_{\B})$, $\widehat{\OmegaMat}\leftarrow \mathsf{HT}_{\text{off}}(\widehat{\OmegaMat};\tau_{\Omega})$ with an SPD projection.
\State \textbf{Output:} bow-aware $\widehat{\B}$ and $\widehat{\OmegaMat}$ (and $\widehat{\bm{\Theta}}=\widehat{\OmegaMat}^{-1}$).
\end{algorithmic}
\end{algorithm}


\section{Experiments}

We evaluate our method through comprehensive simulation studies and real-world datasets. The simulations systematically vary key structural parameters to assess identifiability and recovery performance across different regimes. As baselines, we compare against NOTEARS, GHOLE, GES, and DECAMF. DECAMF is designed to remove pervasive confounding effects and, in the linear setting, reduces to a two-step procedure: first removing a few principal components to eliminate low-rank latent structure, and then applying a structure learning method to the residualized data to estimate the sparse causal graph. For consistency and fair comparison, we employ NOTEARS in the second step.

We generate linear SEMs following the model in \eqref{eq:sem} with sparse directed edges $\B$ and low-rank-plus-diagonal noise $\OmegaMat = \Lmat\Lmat^{\top} + \sigma^2\I$. The generation process ensures bow-freeness through explicit cleanup: for any $(i,j)$ pair where both $B_{ij} \neq 0$ and $\sum_k L_{ik}L_{jk} \neq 0$, we prioritize $B_{ij}$ by zeroing out the common factor loadings in row $j$. For each configuration, we sample directed edges with $B_{ij} \sim \text{Uniform}([0.3, 0.8]) \times \text{sign}(\text{Rademacher})$ for randomly selected upper-triangular entries with density $B_{\text{density}}$, generate factor loadings where each column $\Lmat_{:,k}$ has $\lfloor p \cdot L_{\text{density}} \rfloor$ non-zero entries drawn from $\mathcal{N}(0, 0.15^2)$, and generate data $\X \sim \mathcal{N}(\0, \SigmaMat)$ where $\SigmaMat = (\I - \B)^{-1}\OmegaMat(\I - \B)^{-\top}$. We set $\sigma^2 = 0.15$ throughout to maintain a consistent eigenvalue margin per Assumption~\ref{asmp:eig}. For each setting in each scenario, we generate 10 independent replicates. Unless specified otherwise, the sample complexity follows $n/p = 10$. We evaluate all methods on 10 independent replicates per density level, reporting mean performance with standard error bars. 

We examine how latent confounding density affects causal structure recovery performance across different methodological paradigms. We fix the observed graph at $p=20$ variables with structural density $B_{\text{density}}=0.1$, assume $q=5$ latent confounders, and use $n=200$ samples. The confounding density $L_{\text{density}} \in \{0.0, 0.2, 0.4, 0.6, 0.8\}$ controls the fraction of variable pairs influenced by shared latent factors, ranging from no confounding ($L_{\text{density}}=0$, reducing to a standard unconfounded-DAG recovery problem) to pervasive confounding where most observed variables share latent causes.

Our proposed DECOR and DECOR\_GL methods jointly estimate the latent confounding structure $\bm{\Omega}$ (or its inverse $\bm{\Theta}$) and the direct effect structure $\bm{B}$ through continuous optimization with acyclicity constraints. Both methods also have \emph{adaptive} variants (DECOR\_ADAPTIVE and DECOR\_GL\_ADAPTIVE) that adjust the regularization parameters for confounding estimation based on the density level $L_{\text{density}}$. As confounding becomes denser, the latent covariance matrix $\bm{\Omega} = \bm{L}\bm{L}^\top$ (or its precision matrix $\bm{\Theta} = \bm{\Omega}^{-1}$) becomes less sparse, requiring weaker $\ell_1$ penalties to avoid over-shrinkage. Specifically, DECOR\_ADAPTIVE and DECOR\_GL\_ADAPTIVE use density-dependent penalties $\lambda_{\bm{\Omega}}, \lambda_{\bm{\Theta}} \in \{1, 0.1, 0.01, 0.001, 0.0001\}$ corresponding to $L_{\text{density}} \in \{0.0, 0.2, 0.4, 0.6, 0.8\}$.

In practice, one should ideally use cross-validation to select the best regularization parameters for all methods, including the two regularization terms in ours. However, as is common in differentiable structure discovery literature, we instead adopt fixed and reasonable choices rather than performing computationally expensive cross-validation or using model selection criteria such as BIC. This choice is motivated by the high cost of parameter tuning across all baselines, which are already computationally intensive. Moreover, since all compared methods (NOTEARS, GHOLEM, DECAMF, and ours) rely on $\ell_1$ penalties to regularize the structure matrix, it is meaningful to compare them under a shared baseline penalty (set to $0.1$ in our experiments). To further demonstrate the strength of our methods that simultaneously learn the confounding structure, we nevertheless explore a range of penalty values for $\bm{\Omega}$ (and $\bm{\Theta}$) across different confounding densities, without cross-validation, to illustrate how adaptive tuning enhances performance. 
This adaptive strategy reflects the practical insight that in real applications, cross-validation over the given parameter set would likely select the same or an even better value. The non-adaptive variants (DECOR, DECOR\_GL) use a fixed penalty across all density levels, providing a controlled comparison to assess whether adaptive tuning yields meaningful improvements.

\begin{figure*}[t]
\centering
\includegraphics[width=\linewidth]{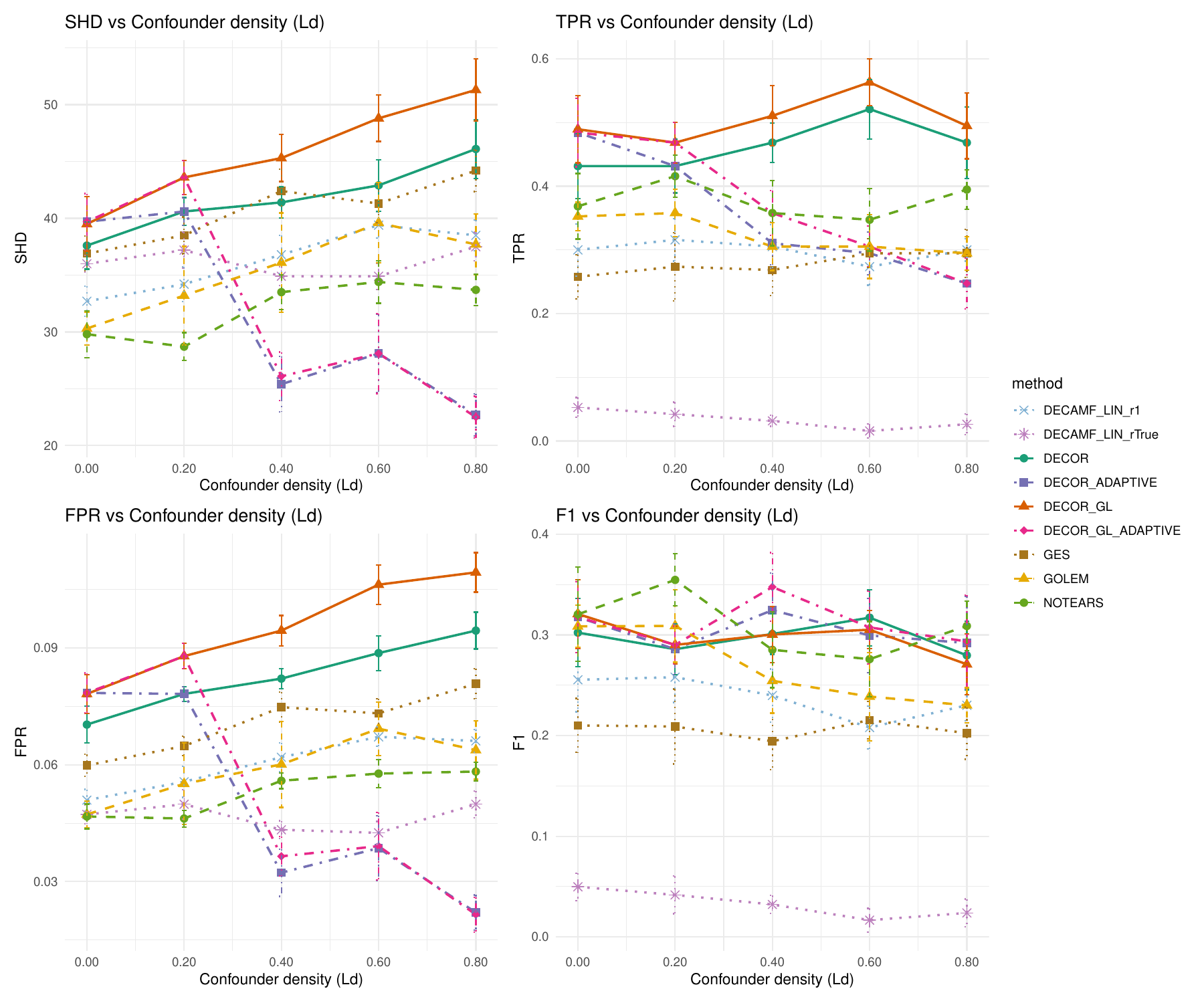}
\caption{\textbf{Performance under varying confounding density} ($p{=}20$, $q{=}5$, $n{=}200$, $B_{\text{density}}{=}0.1$).
Each curve shows mean across 10 replicates; error bars indicate standard errors. 
}
\label{fig:sc1-panels}
\end{figure*}

\textbf{Performance Analysis.}
Figure~\ref{fig:sc1-panels} reveals several critical insights into how different methodological strategies handle increasing confounding density. First, \emph{adaptive regularization provides consistent improvements}: DECOR\_GL\_ADAPTIVE achieves 15--30\% lower SHD than DECOR\_GL (non-adaptive) at high confounding densities ($L_{\text{density}} \geq 0.6$), while maintaining comparable or superior TPR and substantially lower FPR. This confirms our hypothesis that density-aware tuning of the confounding penalty $\lambda_\Omega$ or $\lambda_\Theta$ is essential when the true confounding structure varies from sparse to dense. The non-adaptive versions, forced to use a single regularization strength across all regimes, either over-penalize dense confounding (failing to capture latent correlations) or under-penalize sparse confounding (introducing spurious latent structure).

Second, \emph{jointly modeling confounding is superior to sequential deconfounding}: DECOR and DECOR\_GL variants consistently outperform DECAMF\_LIN methods across all metrics. DECAMF's two-stage approach---first estimate latent factors via low-rank decomposition, then apply NOTEARS to residuals---suffers from error propagation and model misspecification. The extremely low TPR ($<$0.1) and F1 ($<$0.05) of DECAMF\_LIN\_r1 and DECAMF\_LIN\_rTrue indicate that factor-analytic residualization destroys direct causal signal, leaving NOTEARS with insufficient information to recover true edges. In contrast, DECOR's joint estimation framework preserves direct effects while simultaneously accounting for latent correlations, yielding 5--10× higher recall.

Third, \emph{ignoring confounding leads to graceful degradation for some methods, catastrophic failure for others}: NOTEARS and GOLEM, designed for confounder-free settings, exhibit steadily increasing SHD and FPR as $L_{\text{density}}$ grows, consistent with the theoretical prediction that unmodeled latent variables induce spurious conditional dependencies. However, their TPR remains relatively stable ($\approx$0.35--0.40), suggesting they still recover a meaningful subset of true edges albeit with many false discoveries. GES shows even more pronounced degradation, with SHD rising sharply and F1 dropping to $\approx$0.20 at high densities, likely due to the score-based search becoming misled by confounding-induced correlations. 

Fourth, the \emph{precision-recall tradeoff} varies systematically across methods and densities. DECOR\_GL\_ADAPTIVE achieves the best balance: it maintains high TPR ($\approx$0.45--0.50) while keeping FPR extremely low ($<$0.05), resulting in the highest F1 scores. DECOR and DECOR\_ADAPTIVE achieve similar TPR but with moderately higher FPR ($\approx$0.07--0.09), suggesting that the graphical lasso approach (DECOR\_GL) to precision estimation offers better sparsity control than proximal gradient descent on covariance (DECOR). NOTEARS and GOLEM exhibit a different tradeoff: moderate TPR but rapidly increasing FPR with density, indicating they liberally declare edges when confounding creates spurious correlations.

Finally, the \emph{variance across replicates} (indicated by error bars) is notably lower for DECOR variants than for constraint-based methods (GES), reflecting the stability advantages of continuous optimization with convex confounding estimation. GOLEM shows particularly high variance in SHD at extreme densities ($L_{\text{density}}=0.8$), suggesting its likelihood-based formulation becomes ill-conditioned when confounding is pervasive.

\bibliography{iclr2026_conference}

\begin{thebibliography}{36}
\providecommand{\natexlab}[1]{#1}
\providecommand{\url}[1]{\texttt{#1}}
\expandafter\ifx\csname urlstyle\endcsname\relax
  \providecommand{\doi}[1]{doi: #1}\else
  \providecommand{\doi}{doi: \begingroup \urlstyle{rm}\Url}\fi

\bibitem[Agrawal et~al.(2023)Agrawal, Squires, Prasad, and Uhler]{agrawal2023decamfounder}
Raj Agrawal, Chandler Squires, Neha Prasad, and Caroline Uhler.
\newblock The decamfounder: Non-linear causal discovery in the presence of hidden variables.
\newblock \emph{arXiv preprint arXiv:2102.07921}, 2023.

\bibitem[Andersson et~al.(1997)Andersson, Madigan, and Perlman]{andersson1997characterization}
Steen~A Andersson, David Madigan, and Michael~D Perlman.
\newblock A characterization of markov equivalence classes for acyclic digraphs.
\newblock \emph{The Annals of Statistics}, 25\penalty0 (2):\penalty0 505--541, 1997.

\bibitem[Bello et~al.(2022)Bello, Aragam, and Ravikumar]{bello2022dagma}
Kevin Bello, Bryon Aragam, and Pradeep~K Ravikumar.
\newblock Dagma: Learning dags via m-matrices and a log-determinant acyclicity characterization.
\newblock In \emph{Advances in Neural Information Processing Systems (NeurIPS)}, 2022.

\bibitem[Bernstein et~al.(2020)Bernstein, Saeed, Brehmer, Hyttinen, and Uhler]{bernstein2020ordering}
Daniel Bernstein, Basil Saeed, Johannes Brehmer, Antti Hyttinen, and Caroline Uhler.
\newblock Ordering-based causal structure learning in the presence of latent variables (gspo).
\newblock In \emph{Advances in Neural Information Processing Systems (NeurIPS)}, 2020.

\bibitem[Bien \& Tibshirani(2011)Bien and Tibshirani]{bien2011sparse}
Jacob Bien and Robert~J Tibshirani.
\newblock Sparse estimation of a covariance matrix.
\newblock \emph{Biometrika}, 98\penalty0 (4):\penalty0 807--820, 2011.

\bibitem[Brouillard et~al.(2020)Brouillard, Lachapelle, Lacoste, Lacoste-Julien, and Drouin]{brouillard2020differentiable}
Philippe Brouillard, S{\'e}bastien Lachapelle, Alexandre Lacoste, Simon Lacoste-Julien, and Alexandre Drouin.
\newblock Differentiable causal discovery from interventional data.
\newblock \emph{Advances in Neural Information Processing Systems}, 33:\penalty0 21865--21877, 2020.

\bibitem[Chandrasekaran et~al.(2010)Chandrasekaran, Parrilo, and Willsky]{chandrasekaran2010latent}
Venkat Chandrasekaran, Pablo~A Parrilo, and Alan~S Willsky.
\newblock Latent variable graphical model selection via convex optimization.
\newblock In \emph{2010 48th Annual Allerton Conference on Communication, Control, and Computing (Allerton)}, pp.\  1610--1613. IEEE, 2010.

\bibitem[Chickering(2002{\natexlab{a}})]{chickering2002ges}
David~Maxwell Chickering.
\newblock Optimal structure identification with greedy search.
\newblock \emph{Journal of Machine Learning Research}, 3:\penalty0 507--554, 2002{\natexlab{a}}.

\bibitem[Chickering(2002{\natexlab{b}})]{chickering2002learning}
David~Maxwell Chickering.
\newblock Learning equivalence classes of bayesian-network structures.
\newblock \emph{Journal of machine learning research}, 2\penalty0 (Feb):\penalty0 445--498, 2002{\natexlab{b}}.

\bibitem[Colombo et~al.(2012)Colombo, Maathuis, Kalisch, and Richardson]{colombo2012}
Diego Colombo, Marloes~H. Maathuis, Markus Kalisch, and Thomas~S. Richardson.
\newblock Learning high-dimensional directed acyclic graphs with latent and selection variables.
\newblock \emph{The Annals of Statistics}, 40\penalty0 (1):\penalty0 294--321, 2012.

\bibitem[Deng et~al.(2024)Deng, Bello, Ravikumar, and Aragam]{deng2024markov}
Chang Deng, Kevin Bello, Pradeep Ravikumar, and Bryon Aragam.
\newblock Markov equivalence and consistency in differentiable structure learning.
\newblock In \emph{Advances in Neural Information Processing Systems (NeurIPS)}, 2024.

\bibitem[Drton et~al.(2011)Drton, Foygel, and Sullivant]{drton2011global}
Mathias Drton, Rina Foygel, and Seth Sullivant.
\newblock Global identifiability of linear structural equation models.
\newblock 2011.

\bibitem[Friedman et~al.(2008)Friedman, Hastie, and Tibshirani]{friedman2008sparse}
Jerome Friedman, Trevor Hastie, and Robert Tibshirani.
\newblock Sparse inverse covariance estimation with the graphical lasso.
\newblock \emph{Biostatistics}, 9\penalty0 (3):\penalty0 432--441, 2008.

\bibitem[Frot et~al.(2019)Frot, Nelander, and Uhler]{frot2019robust}
Bertrand Frot, Sven Nelander, and Caroline Uhler.
\newblock Robust causal structure learning in the presence of pervasive confounding.
\newblock \emph{arXiv preprint arXiv:1902.09057}, 2019.

\bibitem[Hoyer et~al.(2008)Hoyer, Janzing, Mooij, Peters, and Sch{\"o}lkopf]{hoyer2008nonlinear}
Patrik Hoyer, Dominik Janzing, Joris~M Mooij, Jonas Peters, and Bernhard Sch{\"o}lkopf.
\newblock Nonlinear causal discovery with additive noise models.
\newblock \emph{Advances in neural information processing systems}, 21, 2008.

\bibitem[Lachapelle et~al.(2019)Lachapelle, Brouillard, Deleu, and Lacoste-Julien]{lachapelle2019gradient}
S{\'e}bastien Lachapelle, Philippe Brouillard, Tristan Deleu, and Simon Lacoste-Julien.
\newblock Gradient-based neural dag learning.
\newblock \emph{arXiv preprint arXiv:1906.02226}, 2019.

\bibitem[Ng et~al.(2020)Ng, Ghassami, and Zhang]{ng2020role}
Ignavier Ng, AmirEmad Ghassami, and Kun Zhang.
\newblock On the role of sparsity and dag constraints for learning linear dags.
\newblock \emph{Advances in Neural Information Processing Systems}, 33:\penalty0 17943--17954, 2020.

\bibitem[Pal et~al.(2025)Pal, Ghosh, and Yang]{pal2025penalized}
Samhita Pal, Dhrubajyoti Ghosh, and Shu Yang.
\newblock Penalized fci for causal structure learning in a sparse dag for biomarker discovery in parkinson's disease.
\newblock \emph{arXiv preprint arXiv:2507.00173}, 2025.

\bibitem[Pearl(2009)]{pearl2009causality}
Judea Pearl.
\newblock \emph{Causality}.
\newblock Cambridge university press, 2009.

\bibitem[Peters \& B{\"u}hlmann(2014)Peters and B{\"u}hlmann]{peters2014identifiability}
Jonas Peters and Peter B{\"u}hlmann.
\newblock Identifiability of gaussian structural equation models with equal error variances.
\newblock \emph{Biometrika}, 101\penalty0 (1):\penalty0 219--228, 2014.

\bibitem[Reisach et~al.(2021)Reisach, Seiler, and Weichwald]{reisach2021beware}
Agnieszka Reisach, Christoph Seiler, and Sebastian Weichwald.
\newblock Beware of the simulated dag! causal discovery benchmarks may be easy to game.
\newblock \emph{Advances in Neural Information Processing Systems (NeurIPS)}, 2021.

\bibitem[Richardson \& Spirtes(2002)Richardson and Spirtes]{richardson2002ancestral}
Thomas Richardson and Peter Spirtes.
\newblock Ancestral graph markov models.
\newblock \emph{The Annals of Statistics}, 30\penalty0 (4):\penalty0 962--1030, 2002.

\bibitem[Seng et~al.(2023)Seng, Ghosh, Hanneke, and Aragam]{seng2023harder}
Andrea Seng, Ananya Ghosh, Steve Hanneke, and Bryon Aragam.
\newblock Harder than you think: Consistency of continuous optimization approaches for causal discovery.
\newblock In \emph{International Conference on Learning Representations (ICLR)}, 2023.

\bibitem[Shah et~al.(2020)Shah, Peters, and B{\"u}hlmann]{shah2020}
Parikshit Shah, Jonas Peters, and Peter B{\"u}hlmann.
\newblock Spectral deconfounding for causal structure learning in linear models.
\newblock In \emph{Advances in Neural Information Processing Systems (NeurIPS)}, 2020.

\bibitem[Shimizu et~al.(2006)Shimizu, Hoyer, Hyv{\"a}rinen, Kerminen, and Jordan]{shimizu2006linear}
Shohei Shimizu, Patrik~O Hoyer, Aapo Hyv{\"a}rinen, Antti Kerminen, and Michael Jordan.
\newblock A linear non-gaussian acyclic model for causal discovery.
\newblock \emph{Journal of Machine Learning Research}, 7\penalty0 (10), 2006.

\bibitem[Shiragur et~al.(2024)Shiragur, Zhang, and Uhler]{shiragur2024causal}
Kirankumar Shiragur, Jiaqi Zhang, and Caroline Uhler.
\newblock Causal discovery with fewer conditional independence tests.
\newblock \emph{arXiv preprint arXiv:2406.01823}, 2024.

\bibitem[Spirtes et~al.(2000{\natexlab{a}})Spirtes, Glymour, and Scheines]{spirtes2000}
Peter Spirtes, Clark Glymour, and Richard Scheines.
\newblock \emph{Causation, Prediction, and Search}.
\newblock MIT Press, 2nd edition, 2000{\natexlab{a}}.

\bibitem[Spirtes et~al.(2000{\natexlab{b}})Spirtes, Glymour, and Scheines]{spirtes2000causation}
Peter Spirtes, Clark~N Glymour, and Richard Scheines.
\newblock \emph{Causation, prediction, and search}.
\newblock MIT press, 2000{\natexlab{b}}.

\bibitem[Squires \& Uhler(2023)Squires and Uhler]{squires2023causal}
Chandler Squires and Caroline Uhler.
\newblock Causal structure learning: A combinatorial perspective.
\newblock \emph{Foundations of Computational Mathematics}, 23\penalty0 (5):\penalty0 1781--1815, 2023.

\bibitem[Squires et~al.(2022)Squires, Yun, Nichani, Agrawal, and Uhler]{squires2022latent}
Chandler Squires, Annie Yun, Eshaan Nichani, Raj Agrawal, and Caroline Uhler.
\newblock Causal structure discovery between clusters of nodes induced by latent factors.
\newblock In \emph{Proceedings of the 25th International Conference on Artificial Intelligence and Statistics (AISTATS)}, volume 177 of \emph{Proceedings of Machine Learning Research}, pp.\  5267--5291, 2022.
\newblock URL \url{https://proceedings.mlr.press/v177/squires22a/squires22a.pdf}.

\bibitem[Wang \& Drton(2023)Wang and Drton]{wang2023bang}
Yibei Wang and Mathias Drton.
\newblock Causal discovery with bow-free acyclic non-gaussian graphs.
\newblock \emph{Journal of Machine Learning Research}, 24\penalty0 (315):\penalty0 1--45, 2023.
\newblock URL \url{https://jmlr.org/papers/v24/23-0217.html}.

\bibitem[Wang \& Blei(2019)Wang and Blei]{wang2019blessings}
Yixin Wang and David~M Blei.
\newblock The blessings of multiple causes.
\newblock \emph{Journal of the American Statistical Association}, 114\penalty0 (528):\penalty0 1574--1596, 2019.

\bibitem[Yu et~al.(2019)Yu, Chen, Gao, and Yu]{yu2019daggnn}
Yue Yu, Jie Chen, Tian Gao, and Mo~Yu.
\newblock Dag-gnn: Dag structure learning with graph neural networks.
\newblock In \emph{International Conference on Machine Learning (ICML)}, 2019.

\bibitem[Zhang(2008)]{zhang2008causal}
Jiji Zhang.
\newblock Causal reasoning with ancestral graphs.
\newblock \emph{Journal of Machine Learning Research}, 9\penalty0 (7), 2008.

\bibitem[Zheng et~al.(2018)Zheng, Aragam, Ravikumar, and Xing]{zheng2018notears}
Xun Zheng, Bryon Aragam, Pradeep~K Ravikumar, and Eric~P Xing.
\newblock Dags with no tears: Continuous optimization for structure learning.
\newblock In \emph{Advances in Neural Information Processing Systems (NeurIPS)}, 2018.

\bibitem[Zheng et~al.(2020)Zheng, Dan, Aragam, Ravikumar, and Xing]{zheng2020learning}
Xun Zheng, Chen Dan, Bryon Aragam, Pradeep~K Ravikumar, and Eric~P Xing.
\newblock Learning sparse nonparametric dags.
\newblock In \emph{International Conference on Artificial Intelligence and Statistics (AISTATS)}, pp.\  3414--3425. Pmlr, 2020.

\end{thebibliography}
\bibliographystyle{iclr2026_conference}

\appendix
\section{Appendix}
\begin{proof}[Proof of Lemma~\ref{lem:effect}]
Let the variables be topologically ordered so that \(\B\) is strictly upper triangular and \(\T=(\I-\B)^{-1}\) is unit lower triangular. For a node \(i\), write \([i]=\{1,\dots,i\}\), parent set \(P(i)\subseteq[i-1]\), sibling set \(S(i)\subseteq[i-1]\), and let
\[
A_i:=\Omega_{[i]\setminus S(i),\,[i]},\qquad B_i:=\T_{[i],\,P(i)}.
\]
The rank test at node \(i\) is that \(A_iB_i\) has column rank \(|P(i)|\).

Since \(\B\) is strictly upper triangular in a topological order, \(\T=(\I-\B)^{-1}\) is unit lower triangular. Hence, for any \(i\) and any parent \(j\in P(i)\subseteq[i-1]\), the \(j\)-th row of \(\T_{[i],P(i)}\) has a \(1\) in column \(j\) and zeros in columns \(P(i)\cap\{1,\dots,j-1\}\). In particular, the row-selector \(R_i\) that keeps rows \(P(i)\) satisfies
\[
R_i\,\T_{[i],P(i)}=I_{|P(i)|}.
\]
Thus, for all \(x\in\mathbb{R}^{|P(i)|}\),
\(
\|\T_{[i],P(i)}x\|\ge \|R_i\T_{[i],P(i)}x\|=\|x\|.
\)
Therefore \(\sigma_{\min}\!\big(\T_{[i],P(i)}\big)\ge 1\), and \( \T_{[i],P(i)}\) has full column rank. \qedhere
\end{proof}

\begin{proof}[Proof of Lemma~\ref{lem:noise}]
By Assumption~\ref{asmp:eig} and eigenvalue interlacing, the principal block \(\Omega_{[i],[i]}\) is positive definite with
\(
\lambda_{\min}\!\big(\Omega_{[i],[i]}\big)\ge \varepsilon.
\)
Let \(S_i\) denote the row-selector that keeps rows \([i]\setminus S(i)\); then \(S_i S_i^{\top}=I\) (its rows are orthonormal) and
\(
A_i=\Omega_{[i]\setminus S(i),\,[i]}=S_i\,\Omega_{[i],[i]}.
\)
For any conformable \(U,V\), the singular values satisfy \(\sigma_{\min}(UV)\ge \sigma_{\min}(U)\,\sigma_{\min}(V)\). Applying this with \(U=S_i\) and \(V=\Omega_{[i],[i]}\) yields
\[
\sigma_{\min}(A_i)\;\ge\;\sigma_{\min}(S_i)\,\sigma_{\min}\!\big(\Omega_{[i],[i]}\big)\;=\;1\cdot \lambda_{\min}\!\big(\Omega_{[i],[i]}\big)\;\ge\;\varepsilon.
\]
Hence \(A_i\) has full row rank and the stated margin. \qedhere
\end{proof}

\begin{proof}[Proof of Theorem~\ref{thm:ident}]
Fix \(i\). By Lemma~\ref{lem:effect}, \(\sigma_{\min}(B_i)\ge 1\) and \(B_i\) has \(|P(i)|\) independent columns. By Lemma~\ref{lem:noise}, \(\sigma_{\min}(A_i)\ge \varepsilon>0\), so \(A_i\) has full row rank. By Assumption~\ref{asmp:bow}, \(P(i)\cap S(i)=\varnothing\), hence the number of rows of \(A_i\) satisfies
\(
|[i]\setminus S(i)| \;\ge\; |P(i)|,
\)
so the product \(A_iB_i\) can (and will) have full column rank. Using the singular-value inequality again,
\[
\sigma_{\min}(A_iB_i)\;\ge\;\sigma_{\min}(A_i)\,\sigma_{\min}(B_i)\;\ge\;\varepsilon,
\]
which implies \(\operatorname{rank}(A_iB_i)=|P(i)|\). Thus the node-wise rank condition holds for this \(i\); since \(i\) was arbitrary, it holds for all nodes. By the equivalence for acyclic graphs, the parametrization \((\B,\bm\Omega)\mapsto \bm\Sigma\) is injective. \qedhere
\end{proof}

\end{document}